# Learning Concept Hierarchies from Text Corpora
# using Formal Concept Analysis

**Philipp Cimiano**                                   CIMIANO@AIFB.UNI-KARLSRUHE.DE
*Institute AIFB, University of Karlsruhe*
*Englerstr. 11, 76131 Karlsruhe, Germany*

**Andreas Hotho**                                      HOTHO@CS.UNI-KASSEL.DE
*Knowledge and Data Engineering Group, University of Kassel*
*Wilhelmshöher Allee 73, 34121 Kassel, Germany*

**Steffen Staab**                                      STAAB@UNI-KOBLENZ.DE
*Institute for Computer Science, University of Koblenz-Landau*
*Universitätsstr. 1, 56016 Koblenz, Germany*

## Abstract

We present a novel approach to the automatic acquisition of taxonomies or concept hierarchies from a text corpus. The approach is based on Formal Concept Analysis (FCA), a method mainly used for the analysis of data, i.e. for investigating and processing explicitly given information. We follow Harris' distributional hypothesis and model the context of a certain term as a vector representing syntactic dependencies which are automatically acquired from the text corpus with a linguistic parser. On the basis of this context information, FCA produces a lattice that we convert into a special kind of partial order constituting a concept hierarchy. The approach is evaluated by comparing the resulting concept hierarchies with hand-crafted taxonomies for two domains: tourism and finance. We also directly compare our approach with hierarchical agglomerative clustering as well as with Bi-Section-KMeans as an instance of a divisive clustering algorithm. Furthermore, we investigate the impact of using different measures weighting the contribution of each attribute as well as of applying a particular smoothing technique to cope with data sparseness.

## 1. Introduction

Taxonomies or concept hierarchies are crucial for any knowledge-based system, i.e. a system equipped with declarative knowledge about the domain it deals with and capable of reasoning on the basis of this knowledge. Concept hierarchies are in fact important because they allow to structure information into categories, thus fostering its search and reuse. Further, they allow to formulate rules as well as relations in an abstract and concise way, facilitating the development, refinement and reuse of a knowledge-base. Further, the fact that they allow to generalize over words has shown to provide benefits in a number of applications such as Information Retrieval (Voorhees, 1994) as well as text clustering (Hotho, Staab, & Stumme, 2003) and classification (Bloehdorn & Hotho, 2004). In addition, they also have important applications within Natural Language Processing (e.g. Cimiano, 2003).

However, it is also well known that any knowledge-based system suffers from the so-called *knowledge acquisition bottleneck*, i.e. the difficulty to actually model the domain in question. In order to partially overcome this problem we present a novel approach to automatically learning a concept hierarchy from a text corpus.

Making the knowledge implicitly contained in texts explicit is a great challenge. For example, Brewster, Ciravegna, and Wilks (2003) have argued that text writing and reading is in fact a process of background knowledge maintenance in the sense that basic domain knowledge is assumed, and only the relevant part of knowledge which is the issue of the text or article is mentioned in a more or less explicit way. Actually, knowledge can be found in texts at different levels of explicitness depending on the sort of text considered. Handbooks, textbooks or dictionaries for example contain explicit knowledge in form of definitions such as "a tiger is a mammal" or "mammals such as tigers, lions or elephants". In fact, some researchers have exploited such regular patterns to discover taxonomic or part-of relations in texts (Hearst, 1992; Charniak & Berland, 1999; Iwanska, Mata, & Kruger, 2000; Ahmad, Tariq, Vrusias, & Handy, 2003). However, it seems that the more technical and specialized the texts get, the less basic knowledge we find stated explicitly. Thus, an interesting alternative is to derive knowledge from texts by analyzing how certain terms are used rather than to look for their explicit definition. In these lines the *distributional hypothesis* (Harris, 1968) assumes that terms are similar to the extent to which they share similar linguistic contexts.

In fact, different methods have been proposed in the literature to address the problem of (semi-) automatically deriving a concept hierarchy from text based on the distributional hypothesis. Basically, these methods can be grouped into two classes: the *similarity*-based methods on the one hand and the *set-theoretical* on the other hand. Both methods adopt a vector-space model and represent a word or term as a vector containing features or attributes derived from a certain corpus. There is certainly a great divergence in which attributes are used for this purpose, but typically some sort of syntactic features are used, such as conjunctions, appositions (Caraballo, 1999) or verb-argument dependencies (Hindle, 1990; Pereira, Tishby, & Lee, 1993; Grefenstette, 1994; Faure & Nédellec, 1998).

The first type of methods is characterized by the use of a similarity or distance measure in order to compute the pairwise similarity or distance between vectors corresponding to two words or terms in order to decide if they can be clustered or not. Some prominent examples for this type of method have been developed by Hindle (1990), Pereira et al. (1993), Grefenstette (1994), Faure and Nédellec (1998), Caraballo (1999) as well as Bisson, Nédellec, and Canamero (2000). Set-theoretical approaches partially order the objects according to the inclusion relations between their attribute sets (Petersen, 2002; Sporleder, 2002).

In this paper, we present an approach based on Formal Concept Analysis, a method based on order theory and mainly used for the analysis of data, in particular for discovering inherent relationships between objects described through a set of attributes on the one hand, and the attributes themselves on the other (Ganter & Wille, 1999). In order to derive attributes from a certain corpus, we parse it and extract verb/prepositional phrase (PP)-complement, verb/object and verb/subject dependencies. For each noun appearing as head of these argument positions we then use the corresponding verbs as attributes for building the formal context and then calculating the formal concept lattice on its basis.

Though different methods have been explored in the literature, there is actually a lack of comparative work concerning the task of automatically learning concept hierarchies with clustering techniques. However, as argued by Cimiano, Hotho, and Staab (2004c), ontology engineers need guidelines about the effectiveness, efficiency and trade-offs of different methods in order to decide which techniques to apply in which settings. Thus, we present a comparison along these lines between our FCA-based approach, hierarchical bottom-up (agglomerative) clustering and Bi-Section-KMeans as an instance of a divisive algorithm. In particular, we compare the learned concept hierarchies in terms of similarity with handcrafted reference taxonomies for two domains: tourism and finance. In addition, we examine the impact of using different information measures to weight the significance of a given object/attribute pair. Furthermore, we also investigate the use of a smoothing technique to cope with data sparseness.

The remainder of this paper is organized as follows: Section 2 describes the overall process and Section 3 briefly introduces Formal Concept Analysis and describes the nature of the concept hierarchies we automatically acquire. Section 4 describes the text processing methods we apply to automatically derive context attributes. In Section 5 we discuss in detail our evaluation methodology and present the actual results in Section 6. In particular, we present the comparison of the different approaches as well as the evaluation of the impact of different information measures as well as of our smoothing technique. Before concluding, we discuss some related work in Section 7.

## 2. Overall Process

The overall process of automatically deriving concept hierarchies from text is depicted in Figure 1. First, the corpus is part-of-speech (POS) tagged[1] using TreeTagger (Schmid, 1994) and parsed using LoPar[2] (Schmid, 2000), thus yielding a parse tree for each sentence. Then, verb/subject, verb/object and verb/prepositional phrase dependencies are extracted from these parse trees. In particular, pairs are extracted consisting of the verb and the head of the subject, object or prepositional phrase they subcategorize. Then, the verb and the heads are lemmatized, i.e. assigned to their base form. In order to address data sparseness, the collection of pairs is smoothed, i.e. the frequency of pairs which do not appear in the corpus is estimated on the basis of the frequency of other pairs. The pairs are then weighted according to some statistical measure and only the pairs over a certain threshold are transformed into a formal context to which Formal Concept Analysis is applied. The lattice resulting from this, $(\mathfrak{B}, \leq)$, is transformed into a partial order $(C', \leq')$ which is closer to a concept hierarchy in the traditional sense. As FCA typically leads to a proliferation of concepts, the partial order is compacted in a pruning step, removing abstract concepts and leading to a compacted partial order $(C'', \leq'')$ which is the resulting concept hierarchy. This process is described in detail in Section 3. The process is described more formally by Algorithm 1.

## 3. Formal Concept Analysis

Formal Concept Analysis (FCA) is a method mainly used for the analysis of data, i.e. for deriving implicit relationships between objects described through a set of attributes on the one hand and these attributes on the other. The data are structured into units which are formal abstractions of concepts of human thought, allowing meaningful comprehensible interpretation (Ganter & Wille, 1999). Thus, FCA can be seen as a conceptual clustering technique as it also provides intensional descriptions for the abstract concepts or data units it produces. Central to FCA is the notion of a *formal context*:

---

1. Part-of-speech tagging consists in assigning each word its syntactic category, i.e. noun, verb, adjective etc.
2. http://www.ims.uni-stuttgart.de/projekte/gramotron/SOFTWARE/LoPar-en.html

---

**Algorithm 1** ConstructConceptHierarchy(D,T)

---

/* construct a hierarchy for the terms in $T$ on the basis of the documents in $D$ */

 1: Parses = parse(POS-tag($D$));
 2: SynDeps = tgrep(Parses);
 3: lemmatize(SynDeps);
 4: smooth(SynDeps);
 5: weight(SynDeps);
 6: SynDeps' = applyThreshold(SynDeps);
 7: $K$ = getFormalContext($T$,SynDeps');
 8: $(\mathfrak{B}, \leq)$ = computeLattice($K$);
 9: $(C', \leq')$ = transform($\mathfrak{B}, \leq$);
10: $(C'', \leq'')$ = compact($C', \leq'$);
11: return $(C'', \leq'')$;

---

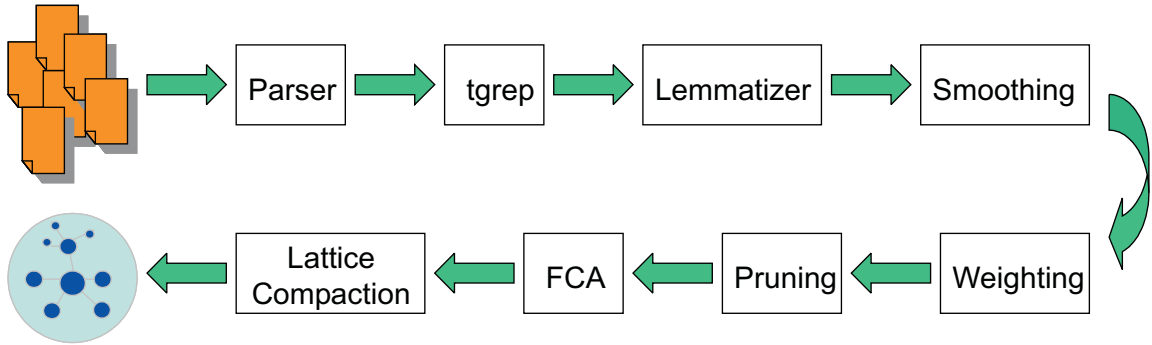

Figure 1: Overall Process

**Definition 1 (Formal Context)**
*A triple (G,M,I) is called a **formal context** if G and M are sets and $I \subseteq G \times M$ is a binary relation between G and M. The elements of G are called **objects**, those of M **attributes** and I is the **incidence** of the context.*

For $A \subseteq G$, we define: $A' := \{m \in M \mid \forall g \in A : (g, m) \in I\}$

and dually for $B \subseteq M$: $B' := \{g \in G \mid \forall m \in B : (g, m) \in I\}$

Intuitively speaking, $A'$ is the set of all attributes common to the objects of $A$, while $B'$ is the set of all objects that have all attributes in $B$. Furthermore, we define what a *formal concept* is:

**Definition 2 (Formal Concept)**
*A pair (A,B) is a **formal concept** of (G,M,I) if and only if $A \subseteq G$, $B \subseteq M$, $A' = B$ and $A = B'$.*

In other words, $(A,B)$ is a **formal concept** if the set of all attributes shared by the objects of $A$ is identical with $B$ and on the other hand $A$ is also the set of all objects that have all attributes in $B$. $A$ is then called the **extent** and $B$ the **intent** of the formal concept $(A,B)$. The formal concepts of a given context are naturally ordered by the **subconcept-superconcept relation** as defined by:

$$(A_1, B_1) \leq (A_2, B_2) \Leftrightarrow A_1 \subseteq A_2 (\Leftrightarrow B_2 \subseteq B_1)$$

Thus, formal concepts are partially ordered with regard to inclusion of their extents or (which is equivalent) inverse inclusion of their intent.

We now give some examples to illustrate our definitions. In the context of the tourism domain one knows for example that things like a *hotel*, an *apartment*, a *car*, a *bike*, a *trip* or an *excursion* can be booked. Furthermore, we know that we can rent a *car*, a *bike* or an *apartment*. Moreover, we can drive a *car* or a *bike*, but only ride a *bike*[3]. In addition, we know that we can join an *excursion* or a *trip*. We can now represent the formal context corresponding to this knowledge as a formal context (see Table 1). The lattice produced by FCA is depicted in Figure 2 (left)[4]. It can be transformed into a special type of concept hierarchy as shown in Figure 2 (right) by removing the bottom element, introducing an ontological concept for each formal concept (named with the intent) and introducing a subconcept for each element in the extent of the formal concept in question.

In order to formally define the transformation of the lattice $(\mathfrak{B}, \leq)$ into the partial order $(C', \leq')$, we assume that the lattice is represented using *reduced labeling*. Reduced labeling as defined in (Ganter & Wille, 1999) means that objects are in the extension of the most specific concept and attributes conversely in the intension of the most general one. This reduced labeling is achieved by introducing functions $\gamma$ and $\mu$. In particular, the name of an object $g$ is attached to the lower half of the corresponding *object concept*, i.e. $\gamma(g) := (\{g\}'', \{g\}')$, while the name of attribute $m$ is located at the upper half of the *attribute concept*, i.e. $\mu(m) := (\{m\}', \{m\}'')$. Now given a lattice $(\mathfrak{B}, \leq)$ of formal concepts for a formal context $K = (G, M, I)$, we transform it into a partial order $(C', \leq')$ as follows:

**Definition 3 (Transformation of $(\mathfrak{B}, \leq)$ to $(C', \leq')$)**
*First of all $C'$ contains objects as well as intents (sets of attributes):*

$$C' := G \cup \{B \mid (A, B) \in \mathfrak{B}\}$$

*Further:*

$$\leq' := \{(g, B_1) \mid \gamma(g) = (A_1, B_1)\} \cup \{(B_1, B_2) \mid (A_1, B_1) \leq (A_2, B_2)\}$$

Finally, as FCA typically produces a high number of concepts, we compress the resulting hierarchy of ontological concepts by removing any inner node whose extension in terms of leave nodes subsumed is the same as the one of its child, i.e. we create a partial order $(C'', \leq''_C)$ as follows:

**Definition 4 (Compacted Concept Hierarchy $(C'', \leq'')$)**
*Assuming that $extension(c)$ is the set of leave nodes dominated by $c$ according to $\leq'_C$:*

$$C'' := \{c_2 \in C' \mid \forall c_1 \in C' \ c_2 \leq'_C c_1 \rightarrow extension(c_2) \neq extension(c_1)\}$$

*Further:*

---

3. According to the Longman Dictionary, in American English it is also possible to *ride* vehicles in general. However, for the purposes of our example we gloss over this fact.

4. The *Concept Explorer* software was used to produce this lattice (see http://sourceforge.net/projects/conexp).

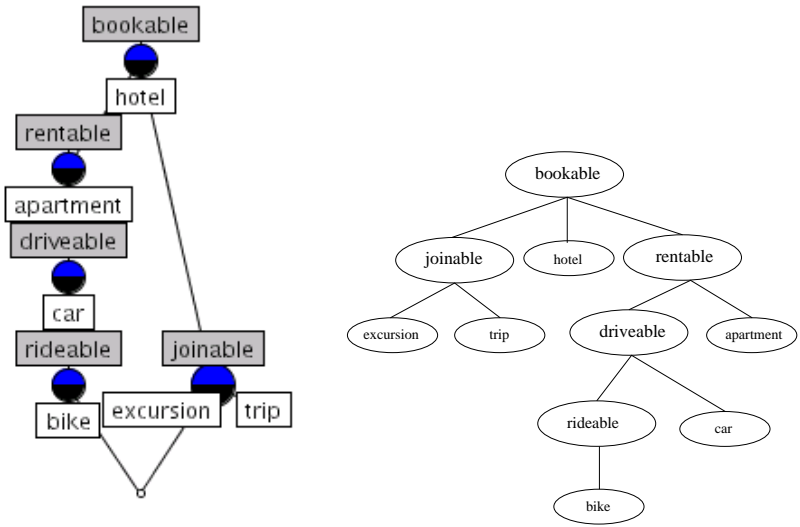

Figure 2: The lattice of formal concepts (left) and the corresponding hierarchy of ontological concepts (right) for the tourism example

$$\leq''_C := \leq'_C \mid_{C'' \times C''}$$

*i.e.* $\leq''_C$ *is the relation* $\leq'_C$ *restricted to pairs of elements of* $C''$.

In particular for the hierarchy in figure 2 (right) we would remove the *rideable* concept.

|           | bookable | rentable | driveable | rideable | joinable |
|-----------|----------|----------|-----------|----------|----------|
| hotel     | x        |          |           |          |          |
| apartment | x        | x        |           |          |          |
| car       | x        | x        | x         |          |          |
| bike      | x        | x        | x         | x        |          |
| excursion | x        |          |           |          | x        |
| trip      | x        |          |           |          | x        |

Table 1: Tourism domain knowledge as formal context

At a first glance, it seems that the hierarchy shown in Figure 2 (right) is somehow odd due to the fact that the labels of abstract concepts are verbs rather than nouns as typically assumed. However, from a formal point of view, concept identifiers have no meaning at all so that we could have just named the concepts with some other arbitrary symbols. The reason why it is handy to introduce 'meaningful' concept identifiers is for the purpose of easier human readability. In fact, if we adopt an extensional interpretation of our hierarchy, we have no problems asserting that the extension of the concept denoted by *bike* is a subset of the extension of the concept of the *rideable* objects in our world. This view is totally compatible with interpreting the concept hierarchy in terms of formal subsumption as given by the logical formula: $\forall x \ (bike(x) \rightarrow rideable(x))$. We thus conclude that from an extensional point of view the 'verb-like' concept identifiers have the same status as any concept label based on a noun. From an intensional point of view, there may not even exist a hypernym with the adequate intension to label a certain abstract concept, such that using a verb-like identifier may even be the most appropriate choice. For example, we could easily replace the identifiers *joinable*, *rideable* and *driveable* by *activity*, *two-wheeled vehicle* and *vehicle*, respectively. However, it is certainly difficult to substitute *rentable* by some 'meaningful' term denoting the same extension, i.e. all the things that can be rented.

It is also important to mention that the learned concept hierarchies represent a conceptualization of a domain with respect to a given corpus in the sense that they represent the relations between terms as they are used in the text. However, corpora represent a very limited view of the world or a certain domain due to the fact that if something is not mentioned, it does not mean that it is not relevant, but simply that it is not an issue for the text in question. This also leads to the fact that certain similarities between terms with respect to the corpus are actually accidental, in the sense that they do not map to a corresponding semantic relation, and which are due to the fact that texts represent an arbitrary snapshot of a domain. Thus, the learned concept hierarchies have to be merely regarded as approximations of the conceptualization of a certain domain.

The task we are now focusing on is: given a certain number of terms referring to concepts relevant for the domain in question, can we derive a concept hierarchy between them? In terms of FCA, the objects are thus given and we need to find the corresponding attributes in order to build an incidence matrix, a lattice and then transform it into a corresponding concept hierarchy. In the following section, we describe how we acquire these attributes automatically from the underlying text collection.

## 4. Text Processing

As already mentioned in the introduction, in order to derive context attributes describing the terms we are interested in, we make use of syntactic dependencies between the verbs appearing in the text collection and the heads of the subject, object and PP-complements they subcategorize. In fact, in previous experiments (Cimiano, Hotho, & Staab, 2004b) we found that using all these dependencies in general leads to better results than any subsets of them. In order to extract these dependencies automatically, we parse the text with LoPar, a trainable, statistical left-corner parser (Schmid, 2000). From the parse trees we then extract the syntactic dependencies between a verb and its subject, object and PP-complement by using tgrep[5]. Finally, we also lemmatize the verbs as well as the head of the subject, object and PP-complement by looking up the lemma in the lexicon provided with LoPar. Lemmatization maps a word to its base form and is in this context used as a sort of normalization of the text. Let's take for instance the following two sentences:

*The museum houses an impressive collection of medieval and modern art. The building combines geometric abstraction with classical references that allude to the Roman influence on the region.*

After parsing these sentences, we would extract the following syntactic dependencies:

---

*houses_subj(museum)*
*houses_obj(collection)*
*combines_subj(building)*
*combines_obj(abstraction)*
*combine_with(references)*
*allude_to(influence)*

By the lemmatization step, *references* is mapped to its base form *reference* and *combines* and *houses* to *combine* and *house*, respectively, such that we yield as a result:

*house_subj(museum)*
*house_obj(collection)*
*combine_subj(building)*
*combine_obj(abstraction)*
*combine_with(reference)*
*allude_to(influence)*

In addition, there are three further important issues to consider:

1. the output of the parser can be erroneous, i.e. not all derived verb/argument dependencies are correct,

2. not all the derived dependencies are 'interesting' in the sense that they will help to discriminate between the different objects,

3. the assumption of completeness of information will never be fulfilled, i.e. the text collection will never be big enough to find all the possible occurrences (compare Zipf, 1932).

To deal with the first two problems, we weight the object/attribute pairs with regard to a certain information measure and only process further those verb/argument relations for which this measure is above some threshold $t$. In particular, we explore the following three information measures (see Cimiano, S.Staab, & Tane, 2003; Cimiano et al., 2004b):

$$Conditional(n, v_{arg}) = P(n|v_{arg}) = \frac{f(n, v_{arg})}{f(v_{arg})}$$

$$PMI(n, v_{arg}) = log\frac{P(n|v_{arg})}{P(n)}$$

$$Resnik(n, v_{arg}) = S_R(v_{arg})\,P(n|v_{arg})$$

where $S_R(v_{arg}) = \sum_{n'} P(n'|v_{arg})\,log\frac{P(n'|v_{arg})}{P(n')}$.

Furthermore, $f(n, v_{arg})$ is the total number of occurrences of a term $n$ as argument *arg* of a verb $v$, $f(v_{arg})$ is the number of occurrences of verb $v$ with such an argument and $P(n)$ is the relative frequency of a term $n$ compared to all other terms. The first information measure is simply the conditional probability of the term $n$ given the argument *arg* of a verb $v$. The second measure $PMI(n, v)$ is the so called *pointwise mutual information* and was used by Hindle (1990) for discovering groups of similar terms. The third measure is inspired by the work of Resnik (1997) and introduces an additional factor $S_R(n, v_{arg})$ which takes into account all the terms appearing in the argument position $arg$ of the verb $v$ in question. In particular, the factor measures the relative entropy of the prior and posterior (considering the verb it appears with) distributions of $n$ and thus the 'selectional strength' of the verb at a given argument position. It is important to mention that in our approach the values of all the above measures are normalized into the interval [0,1].

The third problem requires smoothing of input data. In fact, when working with text corpora, data sparseness is always an issue (Zipf, 1932). A typical method to overcome data sparseness is smoothing (Manning & Schuetze, 1999) which in essence consists in assigning non-zero probabilities to unseen events. For this purpose we apply the technique proposed by Cimiano, Staab, and Tane (2003) in which mutually similar terms are clustered with the result that an occurrence of an attribute with the one term is also counted as an occurrence of that attribute with the other term. As similarity measures we examine the *Cosine*, *Jaccard*, *L1 norm*, *Jensen-Shannon divergence* and *Skew Divergence* measures analyzed and described by Lee (1999):

$$cos(t_1, t_2) = \frac{\sum_{v_{arg} \in V} P(t_1|v_{arg}) P(t_2|v_{arg})}{\sqrt{\sum_{v_{arg} \in V} P(t_1|v_{arg})^2 \sum_{v_{arg} \in V} P(t_2|v_{arg})^2}}$$

$$Jac(t_1, t_2) = \frac{|\{v_{arg}|P(t_1|v_{arg}) > 0 \ and \ P(t_2|v_{arg}) > 0\}|}{|\{v_{arg}|P(t_1|v_{arg}) > 0 \ or \ P(t_2|v_{arg}) > 0\}|}$$

$$L1(t_1, t_2) = \sum_{v_{arg} \in V} |P(t_1|v_{arg}) - P(t_2|v_{arg})|$$

$$JS(t_1, t_2) = \frac{1}{2}[D(P(t_1, V) \ || \ avg(t_1, t_2, V)) + D(P(t_2, V) \ || \ avg(t_1, t_2, V))]$$

$$SD(t_1, t_2) = D(P(t_1, V) \ || \ \alpha \cdot P(t_1, V) \ + (1 - \alpha) \cdot P(t_2, V))$$

where $D(P_1(V) \ || \ P_2(V)) = \sum_{v \in V} P_1(v) \ log \frac{P_1(v)}{P_2(v)}$ and $avg(t_1, t_2, v) = \frac{P(t_1|v) + P(t_2|v)}{2}$

In particular, we implemented these measures using the variants relying only on the elements $v_{arg}$ common to $t_1$ and $t_2$ as described by Lee (1999). Strictly speaking, the Jensen-Shannon as well as the Skew divergences are dissimilarity functions as they measure the average information loss when using one distribution instead of the other. In fact we transform them into similarity measures as $k - f$, where $k$ is a constant and $f$ the dissimilarity function in question. We cluster all the terms which are *mutually similar* with regard to the similarity measure in question, counting more attribute/object pairs than are actually found in the text and thus obtaining also non-zero frequencies for some attribute/object pairs that do not appear literally in the corpus. The overall result is thus a 'smoothing' of the relative frequency landscape by assigning some non-zero relative frequencies to combinations of verbs and objects which were actually not found in the corpus. Here follows the formal definition of mutual similarity:

**Definition 5 (Mutual Similarity)**
*Two terms $n_1$ and $n_2$ are mutually similar iff $n_2 = argmax_{n'} \ sim(n_1, n')$ and $n_1 = argmax_{n'} \ sim(n_2, n')$.*

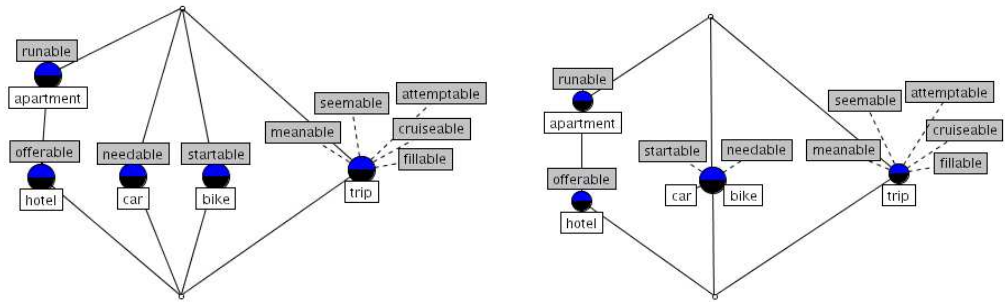

Figure 3: Examples of lattices automatically derived from tourism-related texts without smoothing (left) and with smoothing (right)

According to this definition, two terms $n_1$ and $n_2$ are mutually similar if $n_1$ is the most similar term to $n_2$ with regard to the similarity measure in question and the other way round. Actually, the definition is equivalent to the *reciprocal similarity* of Hindle (1990).

Figure 3 (left) shows an example of a lattice which was automatically derived from a set of texts acquired from *http://www.lonelyplanet.com* as well as *http://www.all-in-all.de*, a web page containing information about the history, accommodation facilities as well as activities of *Mecklenburg Vorpommern*, a region in northeast Germany. We only extracted verb/object pairs for the terms in Table 1 and used the conditional probability to weight the significance of the pairs. For *excursion*, no dependencies were extracted and therefore it was not considered when computing the lattice. The corpus size was about a million words and the threshold used was $t = 0.005$. Assuming that *car* and *bike* are mutually similar, they would be clustered, i.e. *car* would get the attribute *startable* and *bike* the attribute *needable*. The result here is thus the lattice in Figure 3 (right), where *car* and *bike* are in the extension of one and the same concept.

## 5. Evaluation

In order to evaluate our approach we need to assess how good the automatically learned ontologies reflect a given domain. One possibility would be to compute how many of the superconcept relations in the automatically learned ontology are correct. This is for example done by Hearst (1992) or Caraballo (1999). However, due to the fact that our approach, as well as many others (compare Hindle, 1990; Pereira et al., 1993; Grefenstette, 1994), does not produce appropriate names for the abstract concepts generated, it seems difficult to assess the validity of a given superconcept relation. Another possibility is to compute how 'similar' the automatically learned concept hierarchy is with respect to a given hierarchy for the domain in question. Here the crucial question is how to define similarity between concept hierarchies. Though there is a great amount of work in the AI community on how to compute the similarity between trees (Zhang, Statman, & Shasha, 1992; Goddard & Swart, 1996), concept lattices (Belohlavek, 2000), conceptual graphs (Maher, 1993; Myaeng & Lopez-Lopez, 1992) and (plain) graphs (Chartrand, Kubicki, & Schultz, 1998; Zhang, Wang, & Shasha, 1996), it is not clear how these similarity measures also translate to concept hierarchies. An interesting work in these lines is the one presented by Maedche and Staab (2002) in which ontologies are compared along different levels: semiotic, syntactic and pragmatic. In particular, the authors present measures to compare the lexical and taxonomic overlap between two ontologies. Furthermore, they also present an interesting study in which different subjects were asked to model a tourism ontology. The resulting ontologies are compared in terms of the defined similarity measures thus yielding the agreement of different subjects on the task of modeling an ontology.

In order to formally define our evaluation measures, we introduce a *core ontology* model in line with the ontological model presented by Stumme et al. (2003):

**Definition 6 (Core Ontology)**
*A core ontology is a structure $O := (C, root, \leq_C)$ consisting of (i) a set $C$ of concept identifiers, (ii) a designated root element representing the top element of the (iii) partial order $\leq_C$ on $C \cup \{root\}$ such that $\forall c \in C\ c \leq root$, called concept hierarchy or taxonomy.*

For the sake of notational simplicity we adopt the following convention: given an ontology $O_i$, the corresponding set of concepts will be denoted by $C_i$ and the partial order representing the concept hierarchy by $\leq_{C_i}$.

It is important to mention that in the approach presented here, terms are directly identified with concepts, i.e. we neglect the fact that terms can be polysemous.[6] Now, the **Lexical Recall (LR)** of two ontologies $O_1$ and $O_2$ is measured as follows:[7]

$$LR(O_1, O_2) = \frac{|C_1 \cap C_2|}{|C_2|}$$

Take for example the concept hierarchies $O_{auto}$ and $O_{ref}$ depicted in Figure 4. In this example, the Lexical Recall is $LR(O_{auto}, O_{ref}) = \frac{5}{10} = 50\%$.

In order to compare the taxonomy of two ontologies, we use the **Semantic Cotopy (SC)** presented by Maedche and Staab (2002). The Semantic Cotopy of a concept is defined as the set of all its super- and subconcepts:

$$SC(c_i, O_i) := \{c_j \in C_i \mid c_i \leq_C c_j \text{ or } c_j \leq_C c_i\},$$

In what follows we illustrate these and other definitions on the basis of several example concept hierarchies. Take for instance the concept hierarchies in Figure 5. We assume that the left concept hierarchy has been automatically learned with our FCA approach and that the concept hierarchy on the right is a handcrafted one. Further, it is important to point out that the left ontology is, in terms of the arrangement of the leave nodes and abstracting from the labels of the inner nodes, a perfectly learned concept hierarchy. This should thus be reflected by a maximum similarity between both ontologies. The Semantic Cotopy of the concept *vehicle* in the right ontology in Figure 5 is for example {*car, bike, two-wheeled vehicle, vehicle, object-to-rent*} and the Semantic Cotopy of *driveable* in the left ontology is {*bike, car, rideable, driveable, rentable, bookable*}.
It becomes thus already clear that comparing the cotopies of both concepts will not yield the desired results, i.e. a maximum similarity between both concepts. Thus we use a modified version SC' of the Semantic Cotopy in which we only consider the concepts common to both concept hierarchies in the Semantic Cotopy $SC'$ (compare Cimiano et al., 2004b, 2004c), i.e.

---

6. In principle, FCA is able to account for polysemy of terms. However, in this paper we neglect this aspect.

7. As the terms to be ordered hierarchically are given there is no need to measure the lexical precision.

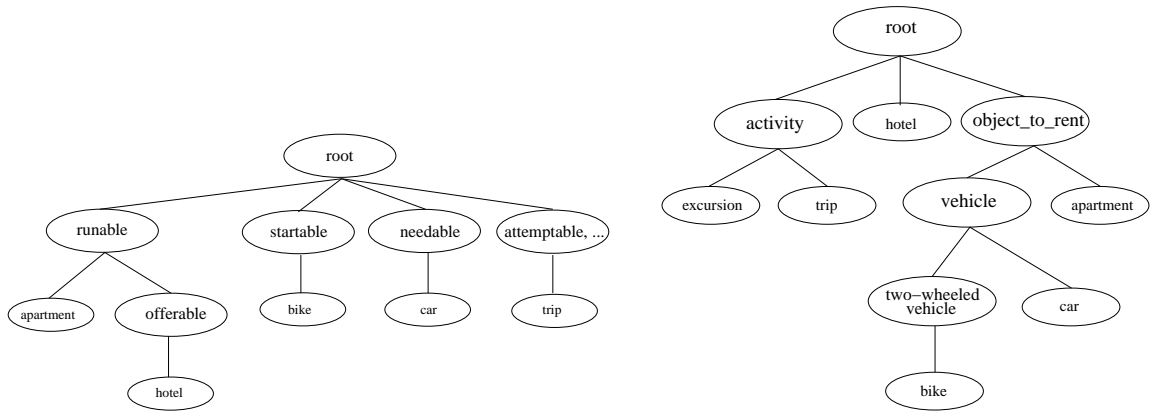

Figure 4: Example for an automatically acquired concept hierarchy $O_{auto}$ (left) compared to the reference concept hierarchy $O_{ref}$ (right)

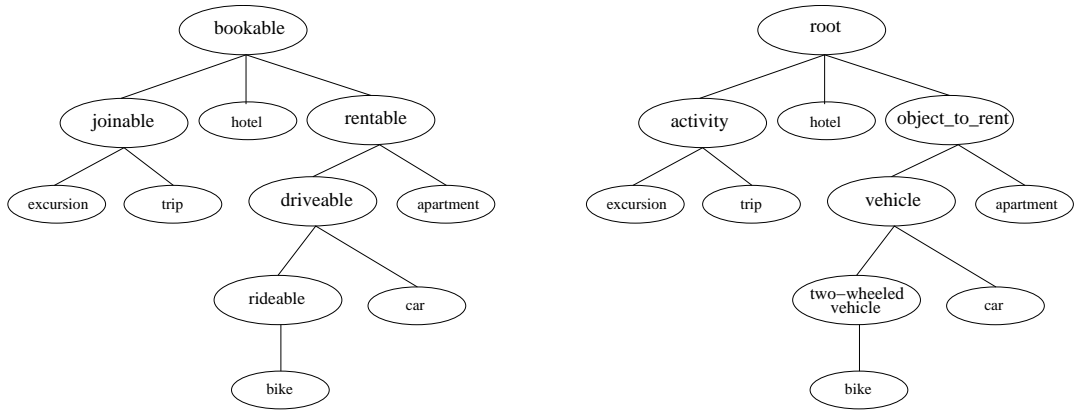

Figure 5: Example for a perfectly learned concept hierarchy $O_{perfect}$ (left) compared to the reference concept hierarchy $O_{ref}$ (right)

$$SC'(c_i, O_1, O_2) := \{c_j \in C_1 \cap C_2 \mid c_j \leq_{C_1} c_i \vee c_i \leq_{C_1} c_j\}$$

By using this **Common Semantic Cotopy** we thus exclude from the comparison concepts such as *runable, offerable, needable, activity, vehicle* etc. which are only in one ontology. So, the Common Semantic Cotopy $SC'$ of the concepts *vehicle* and *driveable* is identical in both ontologies in Figure 5, i.e. {*bike, car*} thus representing a perfect overlap between both concepts, which certainly corresponds to our intuitions about the similarity of both concepts. However, let's now consider the concept hierarchy in Figure 6. The common cotopy of the concept *bike* is {*bike*} in both concept hierarchies. In fact, every leave concept in the left concept hierarchy has a maximum overlap with the corresponding concept in the right ontology. This is certainly undesirable and in fact leads to very high baselines when comparing such trivial concept hierarchies with a reference standard (compare our earlier results Cimiano et al., 2004b, 2004c). Thus, we introduce a further modification of the Semantic Cotopy by excluding the concept itself from its Common Semantic Cotopy, i.e:

$$SC''(c_i, O_1, O_2) := \{c_j \in C_1 \cap C_2 \mid c_j <_{C_1} c_i \vee c_i <_{C_1} c_j\}$$

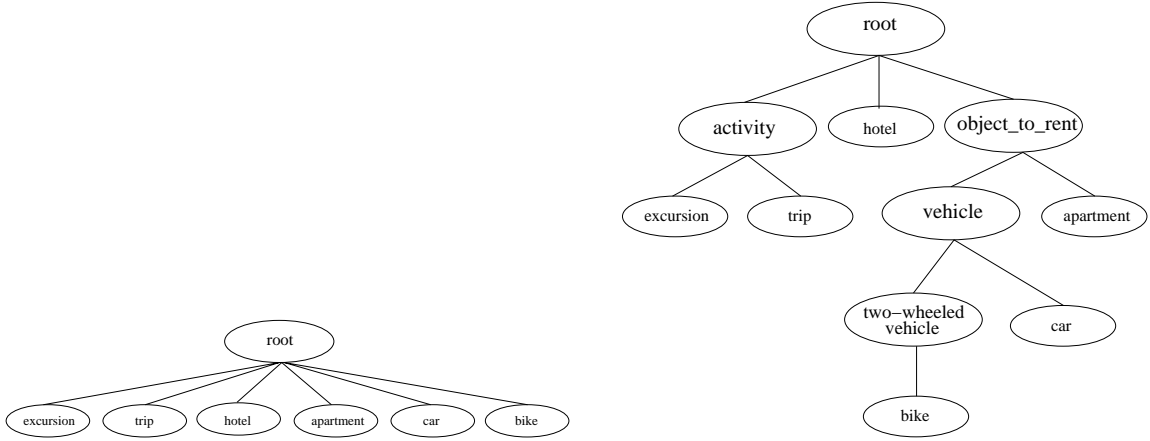

Figure 6: Example for a trivial concept hierarchy $O_{trivial}$ (left) compared to the reference concept hierarchy $O_{ref}$ (right)

This maintains the perfect overlap between *vehicle* and *driveable* in the concept hierarchies in Figure 5, while yielding empty common cotopies for all the leave concepts in the left ontology of Figure 6.

Now, according to Maedche et al. the **Taxonomic Overlap** ($\overline{TO}$) of two ontologies $O_1$ and $O_2$ is computed as follows:

$$\overline{TO}(O_1, O_2) = \frac{1}{|C_1|} \sum_{c \in C_1} TO(c, O_1, O_2)$$

where

$$TO(c, O_1, O_2) := \begin{cases} TO'(c, O_1, O_2) & \text{if } c \in C_2 \\ TO''(c, O_1, O_2) & \text{if } c \notin C_2 \end{cases}$$

and TO' and TO'' are defined as follows:

$$TO'(c, O_1, O_2) := \frac{|SC(c, O_1, O_2) \cap SC(c, O_2, O_1)|}{|SC(c, O_1, O_2) \cup SC(c, O_2, O_1)|}$$

$$TO''(c, O_1, O_2) := max_{c' \in C_2} \frac{|SC(c, O_1, O_2) \cap SC(c', O_2, O_1)|}{|SC(c, O_1, O_2) \cup SC(c', O_2, O_1)|}$$

So, $TO'$ gives the similarity between concepts which are in both ontologies by comparing their respective semantic cotopies. In contrast, $TO''$ gives the similarity between a concept $c \in C_1$ and that concept $c'$ in $C_2$ which maximizes the overlap of the respective semantic cotopies, i.e. it makes an optimistic estimation assuming an overlap that just does not happen to show up at the immediate lexical surface (compare Maedche & Staab, 2002). The Taxonomic Overlap $\overline{TO}(O_1, O_2)$ between the two ontologies is then calculated by averaging over all the taxonomic overlaps of the concepts in $C_1$. In our case it doesn't make sense to calculate the Semantic Cotopy for concepts which are in both ontologies as they represent leave nodes and thus their common semantic cotopies $SC''$ are empty. Thus, we calculate the Taxonomic Overlap between two ontologies as follows:

$$\overline{TO'}(O_1, O_2) = \frac{1}{|C_1 \backslash C_2|} \sum_{c \in C_1 \backslash C_2} max_{c' \in C_2 \cup \{root\}} \frac{|SC''(c, O_1, O_2) \cap SC''(c', O_2, O_1)|}{|SC''(c, O_1, O_2) \cup SC''(c', O_2, O_1)|}$$

Finally, as we do not only want to compute the Taxonomic Overlap in one direction, we introduce the precision, recall and an F-Measure calculating the harmonic mean of both:

$$P(O_1, O_2) = \overline{TO'}(O_1, O_2)$$
$$R(O_1, O_2) = \overline{TO'}(O_2, O_1)$$
$$F(O_1, O_2) = \frac{2 \cdot P(O_1, O_2) \cdot R(O_1, O_2)}{P(O_1, O_2) + R(O_1, O_2)}$$

The importance of balancing recall and precision against each other will be clear in the discussion of a few examples below. Let's consider for example the concept hierarchy $O_{perfect}$ in Figure 5. For the five concepts *bookable, joinable, rentable, driveable* and *rideable* we find a corresponding concept in $O_{ref}$ with a maximum Taxonomic Overlap $TO'$ and the other way round for the concepts *activity, object-to-rent, vehicle* and *two-wheeled-vehicle* in $O_{ref}$, such that $P(O_{perfect}, O_{ref}) = R(O_{perfect}, O_{ref}) = F(O_{perfect}, O_{ref}) = 100\%$.

In the concept hierarchy $O_{\downarrow R}$ shown in Figure 7 the precision is still 100% for the same reasons as above, but due to the fact that the *rideable* concept has been removed there is no corresponding concept for *two-wheeled-vehicle*. The concept maximizing the taxonomic similarity in $O_{ref}$ for *two-wheeled-vehicle* is *driveable* with a Taxonomic Overlap of 0.5. The recall is thus $R(O_{\downarrow R}, O_{ref}) = \overline{TO'}(O_{ref}, O_{\downarrow R}) = \frac{1+1+1+\frac{1}{2}}{4} = 87.5\%$ and the F-Measure decreases to $F(O_{\downarrow R}, O_{ref}) = 93.33\%$.

In the concept hierarchy of $O_{\downarrow P}$ in Figure 8, an additional concept *planable* has been introduced, which reduces the precision to $P(O_{\downarrow P}, O_{ref}) = \frac{1+1+1+1+\frac{1}{2}}{5} = 90\%$, while the recall stays obviously the same at $R(O_{\downarrow P}, O_{ref}) = 100\%$ and thus the F-Measure is $F(O_{\downarrow P}, O_{ref}) = 94.74\%$. It becomes thus clear why it is important to measure the precision and recall of the automatically learned concept hierarchies and balance them against each other by the harmonic mean or F-Measure. For the automatically learned concept hierarchy $O_{auto}$ in Figure 4 the precision is $P(O_{auto}, O_{ref}) = \frac{\frac{2}{6}+\frac{1}{6}+1+\frac{1}{2}+\frac{1}{2}}{5} = 50\%$, the recall $R(O_{auto}, O_{ref}) = \frac{\frac{1}{2}+\frac{3}{5}+\frac{2}{5}+1}{4} = 62.5\%$ and thus the F-Measure $F(O_{auto}, O_{ref}) = 55.56\%$.

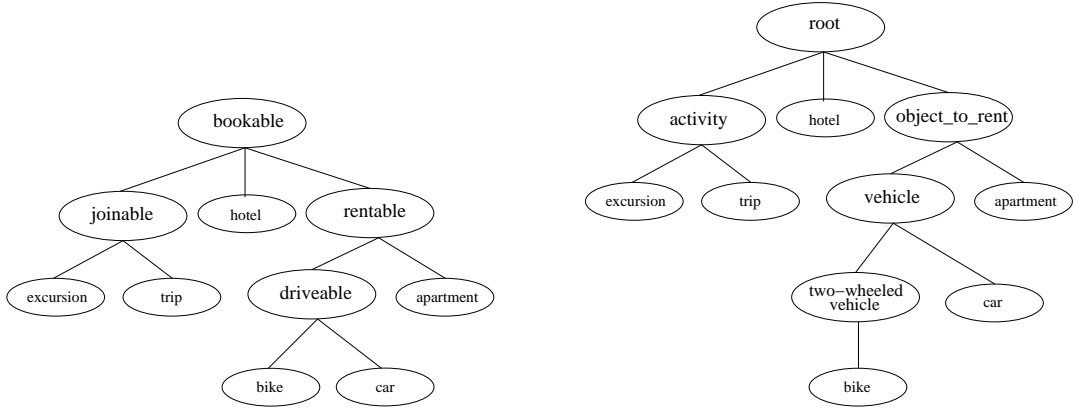

Figure 7: Example for a concept hierarchy with lower recall ($O_{\downarrow R}$) compared to the reference concept hierarchy $O_{ref}$

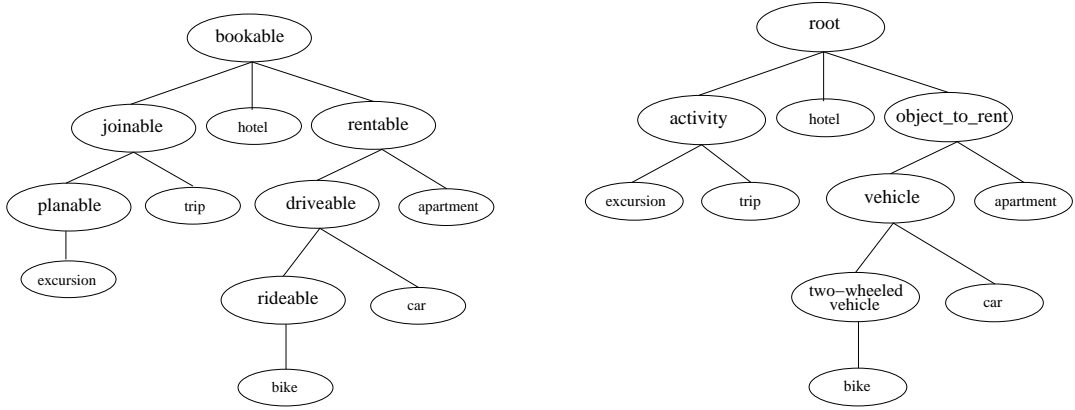

Figure 8: Example for a concept hierarchy with lower precision ($O_{\downarrow P}$) compared to reference concept hierarchy $O_{ref}$

As a comparison, for the trivial concept hierarchy $O_{trivial}$ in Figure 6 we get $P(O_{trivial}, O_{ref}) = 100\%$ (per definition), $R(O_{trivial}, O_{ref}) = \frac{\frac{2}{6} + \frac{3}{6} + \frac{2}{6} + \frac{1}{6}}{4} = 33.33\%$ and $F(O_{trivial}, O_{ref}) = 50\%$.

It is important to mention that though in our toy examples the difference with respect to these measures between the automatically learned concept hierarchy $O_{auto}$ and the trivial concept hierarchy $O_{trivial}$ is not so big, when considering real-world concept hierarchies with a much higher number of concepts it is clear that the F-Measures for trivial concept hierarchies will be very low (see the results in Section 6).

Finally, we also calculate the harmonic mean of the lexical recall and the F-Measure as follows:

$$F'(O_1, O_2) = \frac{2 \cdot LR(O_1, O_2) \cdot F(O_1, O_2)}{LR(O_1, O_2) + F(O_1, O_2)}$$

|               | Tourism | Finance |
|---------------|---------|---------|
| No. Concepts  | 293     | 1223    |
| No. Leaves    | 236     | 861     |
| Avg. Depth    | 3.99    | 4.57    |
| Max. Depth    | 6       | 13      |
| Max. Children | 21      | 33      |
| Avg. Children | 5.26    | 3.5     |

Table 2: Ontology statistics

For the automatically learned concept hierarchy $O_{auto}$, we get for example:

$$F'(O_1, O_2) = \frac{2 \cdot 50\% \cdot 55.56\%}{50\% + 55.56\%} = 52.63\%.$$

## 6. Results

As already mentioned above, we evaluate our approach on two domains: tourism and finance. The ontology for the tourism domain is the reference ontology of the comparison study presented by Maedche and Staab (2002), which was modeled by an experienced ontology engineer. The finance ontology is basically the one developed within the GETESS project (Staab et al., 1999); it was designed for the purpose of analyzing German texts on the Web, but also English labels are available for many of the concepts. Moreover, we manually added the English labels for those concepts whose German label has an English counterpart with the result that most of the concepts (>95%) finally yielded also an English label.[8] The tourism domain ontology consists of 293 concepts, while the finance domain ontology is bigger with a total of 1223 concepts[9]. Table 2 summarizes some facts about the concept hierarchies of the ontologies, such as the total number of concepts, the total number of leave concepts, the average and maximal length of the paths from a leave to the root node as well as the average and maximal number of children of a concept (without considering leave concepts).

As domain-specific text collection for the tourism domain we use texts acquired from the above mentioned web sites, i.e. from *http://www.lonelyplanet.com* as well as from *http://www.all-in-all.de*. Furthermore, we also used a general corpus, the British National Corpus[10]. Altogether, the corpus size was over 118 Million tokens. For the finance domain we considered Reuters news from 1987 with over 185 Million tokens[11].

### 6.1 Comparison

The best F-Measure for the tourism dataset is $F_{FCA,tourism} = 40.52\%$ (at a threshold of $t = 0.005$), corresponding to a precision of $P_{FCA,tourism} = 29.33\%$ and a recall of $R_{FCA,tourism} = 65.49\%$.

---

8. There were some concepts which did not have a direct counterpart in the other language.

9. The ontologies can be downloaded at http://www.aifb.uni-karlsruhe.de/WBS/pci/TourismGoldStandard.isa and http://www.aifb.uni-karlsruhe.de/WBS/pci/FinanceGoldStandard.isa, respectively

10. http://www.natcorp.ox.ac.uk/

11. http://www.daviddlewis.com/resources/testcollections/reuters21578/

For the finance dataset, the corresponding values are $F_{FCA,finance} = 33.11\%$, $P_{FCA,finance} = 29.93\%$ and $R_{FCA,finance} = 37.05\%$.

The Lexical Recall obviously also decreases with increasing threshold $t$ such that overall the F-Measure $F'$ also decreases inverse proportionally to $t$. Overall, the best results in terms of F' are $F'_{FCA,tourism} = 44.69\%$ for the tourism dataset and $F'_{FCA,finance} = 38.85\%$ for the finance dataset. The reason that the results on the finance dataset are slightly lower is probably due to the more technical nature of the domain (compared to the tourism domain) and also to the fact that the concept hierarchy to be learned is bigger.

In order to evaluate our FCA-based approach, we compare it with hierarchical agglomerative clustering and Bi-Section-KMeans. Hierarchical agglomerative clustering (compare Duda, Hart, & Stork, 2001) is a similarity-based bottom-up clustering technique in which at the beginning every term forms a cluster of its own. Then the algorithm iterates over the step that merges the two most similar clusters still available, until one arrives at a universal cluster that contains all the terms.

In our experiments, we use three different strategies to calculate the similarity between clusters: *complete*, *average* and *single*-linkage. The three strategies may be based on the same similarity measure between terms, i.e. the cosine measure in our experiments, but they measure the similarity between two non-trivial clusters in different ways.

*Single linkage* defines the similarity between two clusters $P$ and $Q$ as $\max_{p \in P, q \in Q} sim(p, q)$, considering the closest pair between the two clusters. *Complete* linkage considers the two most dissimilar terms, i.e. $\min_{p \in P, q \in Q} sim(p, q)$. Finally, *average-linkage* computes the average similarity of the terms of the two clusters, i.e. $\frac{1}{|P||Q|} \sum_{p \in P, q \in Q} sim(p, q)$. The reader should note that we prohibit the merging of clusters with similarity 0 and rather order them under a fictive universal cluster 'root'. This corresponds exactly to the way FCA creates and orders objects with no attributes in common. The time complexity of a naive implementation of agglomerative clustering is $O(n^3)$, while efficient implementations have a worst-time complexity of $O(n^2 \log n)$ for complete linkage as it requires sorting of the similarity matrix (Day & Edelsbrunner, 1984), $O(n^2)$ for average linkage if the vectors are length-normalized and the similarity measure is the cosine (see Manning & Schuetze, 1999) and $O(n^2)$ for single linkage (compare Sibson, 1973).[12]

Bi-Section-KMeans is defined as an outer loop around standard KMeans (Steinbach, Karypis, & Kumar, 2000). In order to generate $k$ clusters, Bi-Section-KMeans repeatedly applies KMeans. Bi-Section-KMeans is initiated with the universal cluster containing all terms. Then it loops: It selects the cluster with the largest variance[13] and it calls KMeans in order to split this cluster into exactly two subclusters. The loop is repeated $k-1$ times such that $k$ non-overlapping subclusters are generated. As similarity measure we also use the cosine measure. The complexity of Bi-Section-KMeans is $O(k \cdot n)$. As we want to generate a complete cluster tree with $n$ clusters the complexity is thus $O(n^2)$. Furthermore, as Bi-Section-KMeans is a randomized algorithm, we produce ten runs and average the obtained results.

We compare the different approaches along the lines of the measures described in Section 5. Figure 9 shows the results in terms of F-Measure $F$ over Lexical Recall for both domains and all the clustering approaches. In particular, it shows 8 data points corresponding to the thresholds 0.005, 0.01, 0.05, 0.1, 0.3, 0.5, 0.7 and 0.9. First of all it seems important to discuss the baselines for our approach. The baselines for our approach are the trivial concept hierarchies which are generated when no objects have attributes in common. Such trivial concept hierarchies are generated from threshold 0.7 on our datasets and by definition have a precision of 100% and a recall close to 0. While the baselines for FCA and the agglomerative clustering algorithm are the same, Bi-Section-KMeans is producing a hierarchy by random binary splits which results in higher F' values. These trivial hierarchies represent an absolute baseline in the sense that no algorithm could perform worse. It can also be seen in Figure 9 that our FCA-based approach performs better than the other approaches on both domains. As can be observed in Figure 10, showing recall over precision, the main reason for this is that the FCA-based approach yields a higher recall than the other a approaches, while maintaining the precision at reasonable levels.

---

12. See also http://www-csli.stanford.edu/~schuetze/completelink.html on this topic.

13. Though we don't make use of it in our experiments, it is also possible to select the largest cluster for splitting.

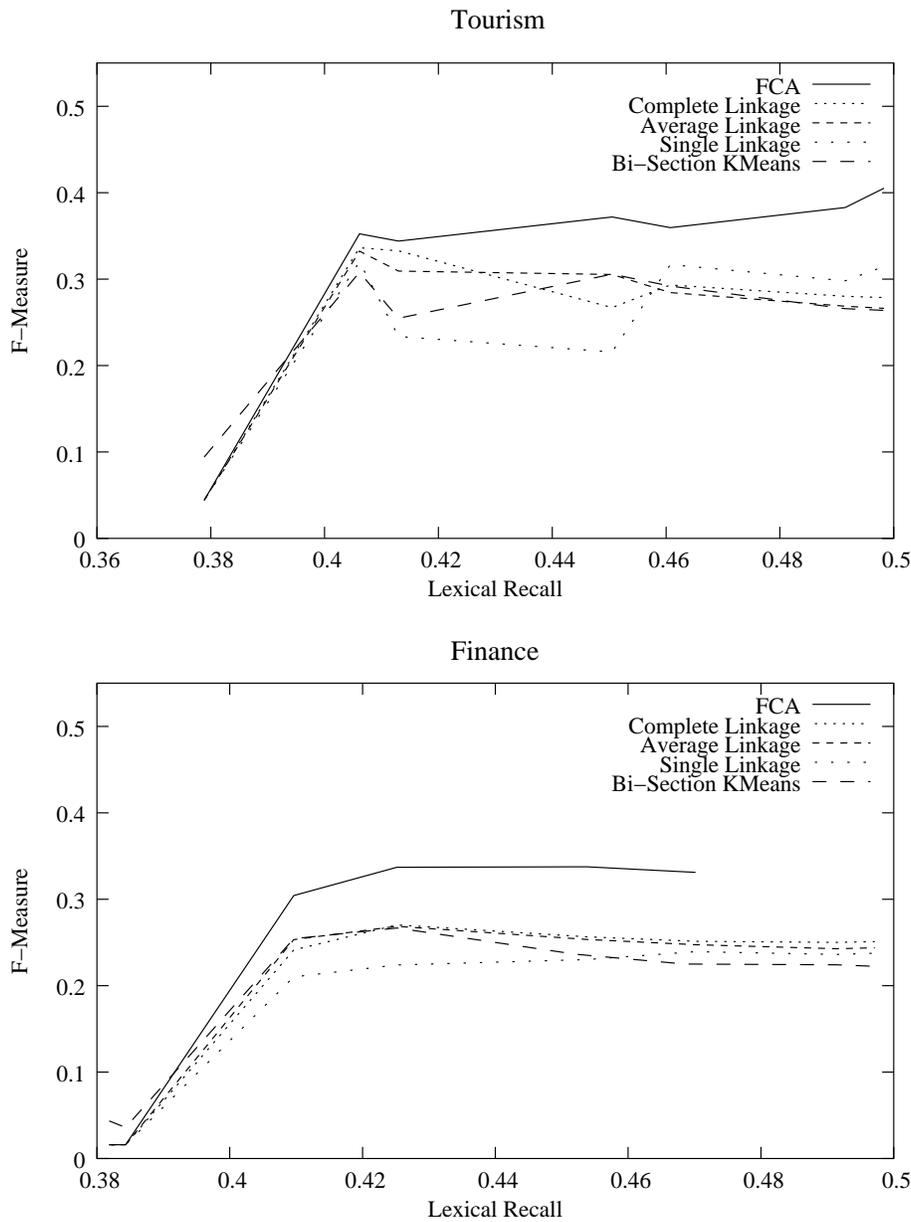

Figure 9: Results for the FCA-based approach: F-Measure over Lexical Recall for the tourism and finance domains

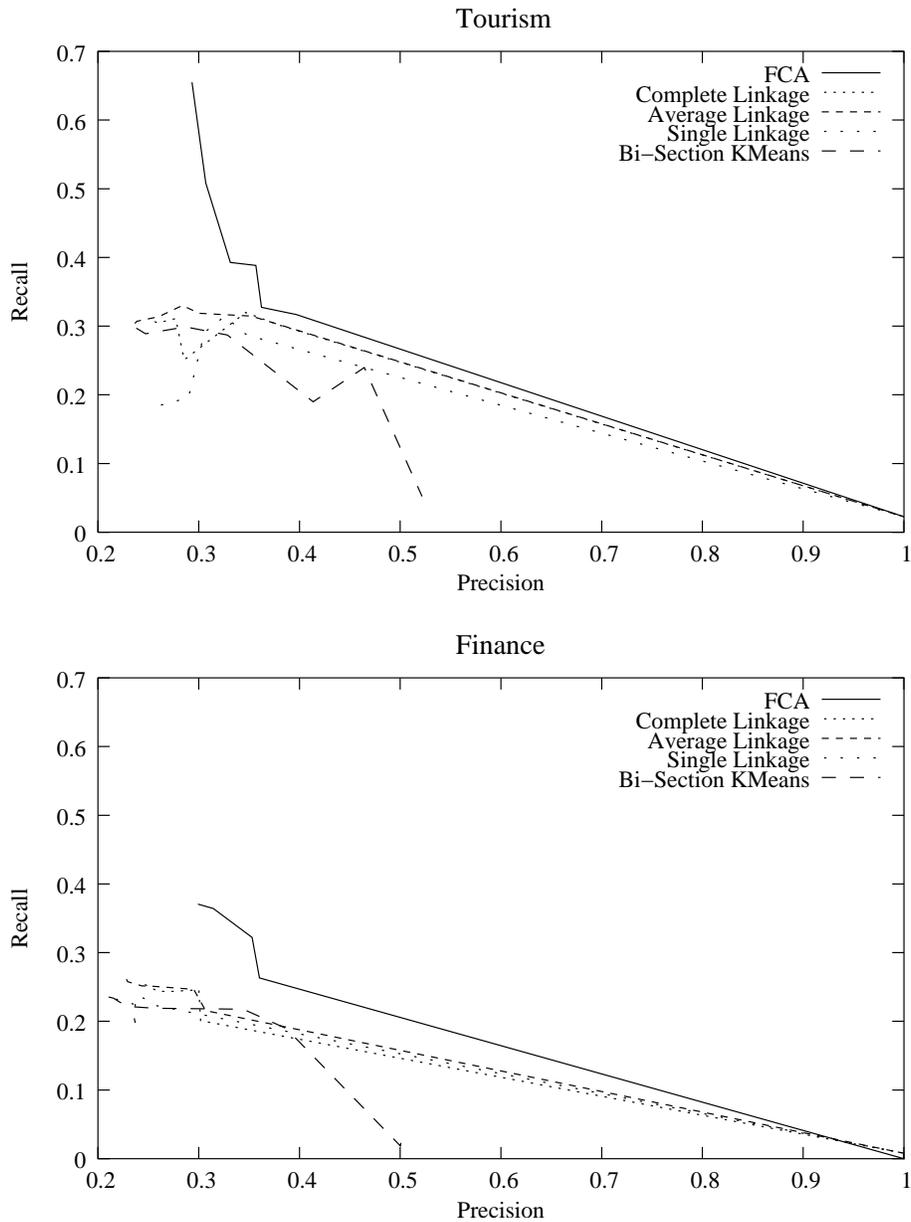

Figure 10: Results for the FCA-based approach: Recall over precision for the tourism and finance domains

On the tourism domain, the second best result is achieved by the agglomerative algorithm with the single-linkage strategy, followed by the ones with average-linkage and complete-linkage (in this order), while the worst results are obtained when using Bi-Section-KMeans (compare Table 3). On the finance domain, the second best results are achieved by the agglomerative algorithm with the complete-linkage strategy followed by the one with the average-linkage strategy, Bi-Section-KMeans and the one with the single-linkage strategy (in this order). Overall, it is valid to claim that FCA outperforms the other clustering algorithms on both datasets. Having a closer look at Table 3, the reason becomes clear, i.e. FCA has a much higher recall than the other approaches, while the precision is more or less comparable. This is due to the fact that FCA generates a higher number of concepts than the other clustering algorithms thus increasing the recall. Interestingly, at the same time the precision of these concepts remains reasonably high thus also yielding higher F-Measures $F$ and $F'$.

| | Tourism | | | | Finance | | | |
|---|---|---|---|---|---|---|---|---|
| | P | R | F | F' | P | R | F | F' |
| FCA | 29.33% | **65.49%** | **40.52%** | **44.69%** | 29.93% | **37.05%** | **33.11%** | **38.85%** |
| Complete Link | 34.67% | 31.98% | 33.27% | 36.85% | 24.56% | 25.65% | 25.09% | 33.35% |
| Average Link | **35.21%** | 31.45% | 33.23% | 36.55% | 29.51% | 24.65% | 26.86% | 32.92% |
| Single Link | 34.78% | 28.71% | 31.46% | 38.57% | 25.23% | 22.44% | 23..75% | 32.15% |
| Bi-Sec. KMeans | 32.85% | 28.71% | 30.64% | 36.42% | **34.41%** | 21.77% | 26.67% | 32.77% |

Table 3: Results of the comparison of different clustering approaches

An interesting question is thus how big the produced concept hierarchies are. Figure 11 shows the size of the concept hierarchies in terms of number of concepts over the threshold parameter $t$ for the different approaches on both domains. It is important to explain why the number of concepts is different for the different agglomerative algorithms as well as Bi-Section-KMeans as in principle the size should always be $2 \cdot n$, where $n$ is the number of objects to be clustered. However, as objects with no similarity to other objects are added directly under the fictive root element, the size of the concept hierarchies varies depending on the way the similarities are calculated. In general, the sizes of the agglomerative and divisive approaches are similar, while at lower thresholds FCA yields concept hierarchies with much higher number of concepts. From threshold 0.3 on, the sizes of the hierarchies produced by all the different approaches are quite similar. Table 4 shows the results for all approaches using the thresholds 0.3 and 0.5. In particular we can conclude that FCA also outperforms the other approaches on both domains when producing a similar number of concepts.

In general, we have not determined the statistical significance of the results presented in this paper as FCA, in contrast to Bi-Section-K-Means, is a deterministic algorithm which does not depend on any random seeding. Our implementation of the agglomerative clustering algorithm is also deterministic given a certain order of the terms to be clustered. Thus, the only possibility to calculate the significance of our results would be to produce different runs by randomly leaving out parts of the corpus and calculating a statistical significance over the different runs. We have not pursued this direction further as the fact that FCA performs better in our setting is clear from the results in Table 3.

|  | Tourism | | Finance | |
|---|---|---|---|---|
| Threshold | 0.3 | 0.5 | 0.3 | 0.5 |
| FCA | **37.53%** | **37.74%** | **37.59%** | **34.92%** |
| Complete Link | 36.85% | 36.78% | 33.05% | 30.37% |
| Single Link | 29.84% | 35.79% | 29.34% | 27.79% |
| Average Link | 35.36% | 36.55% | 32.92% | 31.30% |
| Bi-Sec. KMeans | 31.50% | 35.02% | 32.77% | 31.38% |

Table 4: Comparison of results at thresholds 0.3 and 0.5 in terms of F'

|  | Conditional | PMI | Resnik |
|---|---|---|---|
| FCA | | | |
| Tourism | **44.69%** | 44.51% | 43.31% |
| Finance | 38.85% | **38.96%** | 38.87 % |
| Complete Linkage | | | |
| Tourism | **36.85%** | 27.56% | 23.52% |
| Finance | **33.35%** | 22.29% | 22.96% |
| Average Linkage | | | |
| Tourism | **36.55%** | 26.90% | 23.93% |
| Finance | **32.92%** | 23.78% | 23.26% |
| Single Linkage | | | |
| Tourism | **38.57%** | 30.73% | 28.63% |
| Finance | **32.15%** | 25.47% | 23.46% |
| Bi-Section-KMeans | | | |
| Tourism | **36.42%** | 27.32% | 29.33% |
| Finance | **32.77%** | 26.52% | 24.00% |

Table 5: Comparison of results for different information measures in terms of F'

## 6.2 Information Measures

As already anticipated in Section 4, the different information measures are also subject of our analysis. Table 5 presents the best results for the different clustering approaches and information measures. It can be concluded from these results that using the *PMI* or *Resnik* measures produces worse results on the tourism dataset, while yielding only slightly better results on the finance dataset for the FCA-based approach. It is also interesting to observe that compared to the FCA-based approach, the other clustering approaches are much more sensitive to the information measure used. Overall, the use of the *Conditional* information measure seems a reasonable choice.

## 6.3 Smoothing

We applied our smoothing method described in section 4 to both datasets in order to find out in how far the clustering of terms improves the results of the FCA-based approach. As information measure we use in this experiment the conditional probability as it performs reasonably well as shown in Section 6.2. In particular we used the following similarity measures: the cosine measure, the Jaccard coefficient, the L1 norm as well as the Jensen-Shannon and the Skew divergences (compare Lee, 1999). Table 6 shows the impact of this smoothing technique in terms of the number of object/attribute terms added to the dataset. The *Skew Divergence* is excluded because it did not yield any mutually similar terms. It can be observed that smoothing by mutual similarity based on the cosine measure produces the most previously unseen object/attribute pairs, followed by the Jaccard, L1 and Jensen-Shannon divergence (in this order). Table 7 shows the results for the different similarity measures. The tables in appendix A list the mutually similar terms for the different domains and similarity measures. The results show that our smoothing technique actually yields worse results on both domains and for all similarity measures used.

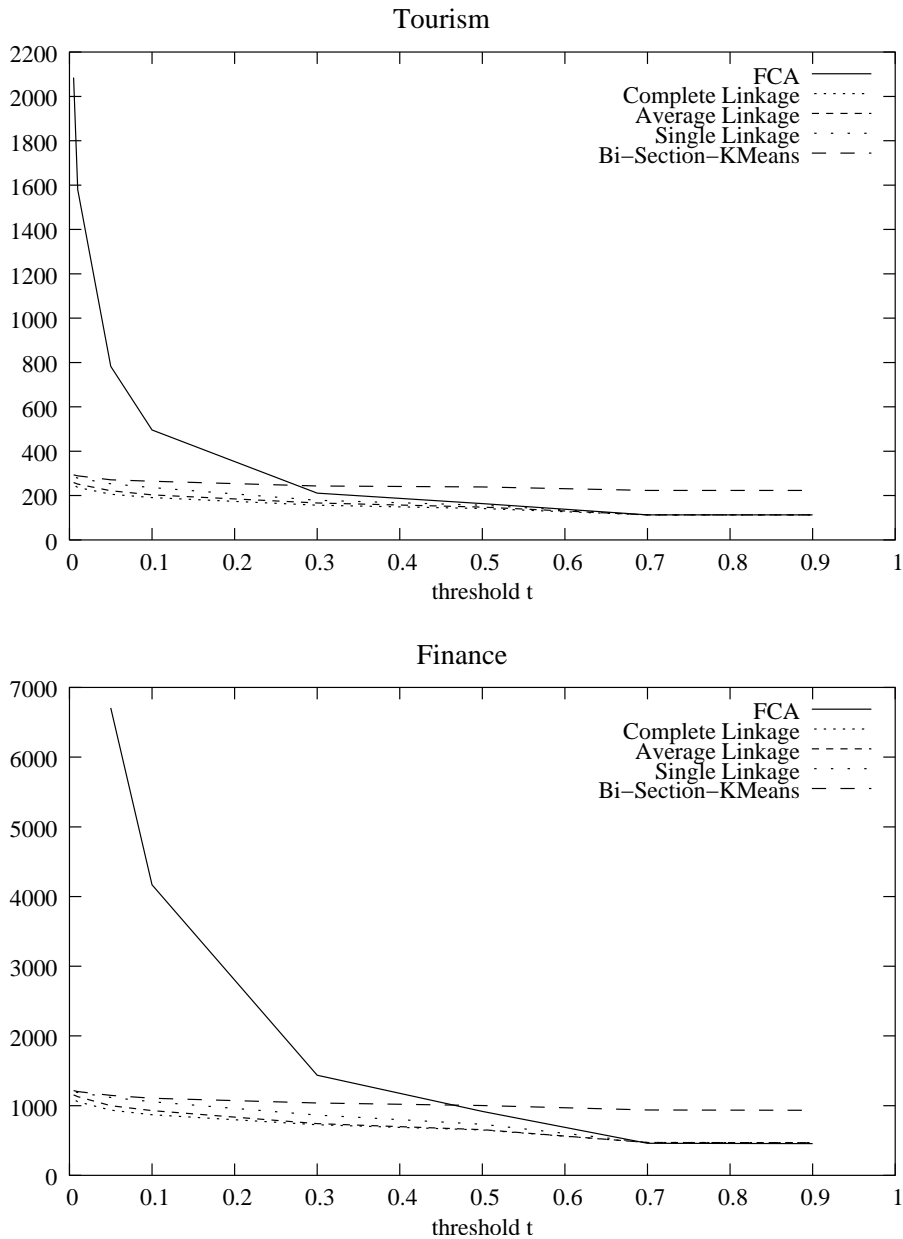

Figure 11: Sizes of concept hierarchies for the different approaches on the tourism and finance domains: number of concepts over threshold $t$

| | Baseline | Jaccard | Cosine | L1 | JS |
|---|---|---|---|---|---|
| Tourism | 525912 | 531041 (+ 5129) | 534709 (+ 8797) | 530695 (+ 4783) | 528892 (+ 2980) |
| Finance | 577607 | 599691 (+ 22084) | 634954 (+ 57347) | 584821 (+ 7214) | 583526 (+ 5919) |

Table 6: Impact of Smoothing Technique in terms of new object/attribute pairs

| | Baseline | Jaccard | Cosine | L1 | JS |
|---|---|---|---|---|---|
| Tourism | **44.69%** | 39.54% | 41.81% | 41.59% | 42.35% |
| Finance | **38.85%** | 38.63% | 36.69% | 38.48% | 38.66% |

Table 7: Results of Smoothing in terms of F-Measure F'

## 6.4 Discussion

We have shown that our FCA-based approach is a reasonable alternative to similarity-based clustering approaches, even yielding better results on our datasets with regard to the $F'$ measure defined in Section 5. The main reason for this is that the concept hierarchies produced by FCA yield a higher recall due to the higher number of concepts, while maintaining the precision relatively high at the same time. Furthermore, we have shown that the conditional probability performs reasonably well as information measure compared to other more elaborate measures such as PMI or the one used by Resnik (1997). Unfortunately, applying a smoothing method based on clustering mutually similar terms does not improve the quality of the automatically learned concept hierarchies. Table 8 highlights the fact that every approach has its own benefits and drawbacks. The main benefit of using FCA is on the one hand that on our datasets it performed better than the other algorithms thus producing better concept hierarchies On the other hand, it does not only generate clusters - formal concepts to be more specific - but it also provides an intensional description for these clusters thus contributing to better understanding by the ontology engineer (compare Figure 2 (left)). In contrast, similarity-based methods do not provide the same level of traceability due to the fact that it is the numerical value of the similarity between two high-dimensional vectors which drives the clustering process and which thus remains opaque to the engineer. The agglomerative and divisive approach are different in this respect as in the agglomerative paradigm, initial merges of small-size clusters correspond to high degrees of similarity and are thus more understandable, while in the divisive paradigm the splitting of clusters aims at minimizing the overall cluster variance thus being harder to trace.

A clear disadvantage of FCA is that the size of the lattice can get exponential in the size of the context in the worst case thus resulting in an exponential time complexity — compared to $O(n^2 \log n)$ and $O(n^2)$ for agglomerative clustering and Bi-Section-KMeans, respectively. The implementation of FCA we have used is the *concepts* tool by Christian Lindig[14], which basically implements Ganter's Next Closure algorithm (Ganter & Reuter, 1991; Ganter & Wille, 1999) with the extension of Aloui for computing the covering relation as described by (Godin, Missaoui, & Alaoui, 1995). Figure 12 shows the number of seconds over the number of attribute/object pairs it took FCA to compute the lattice of formal concepts compared to the time needed by a naive $O(n^3)$ implementation of the agglomerative algorithm with complete linkage. It can be seen that FCA performs quite efficiently compared to the agglomerative clustering algorithm. This is due to the fact that the object/attribute matrix is sparsely populated. Such observations have already been made before. Godin et al. (1995) for example suspect that the lattice size linearly increases with the number of attributes per object. Lindig (2000) presents empirical results analyzing contexts with a fill ratio below 0.1 and comes to the conclusion that the lattice size grows quadratically with respect to the size of the incidence relation $I$. Similar findings are also reported by Carpineto and Romano (1996).

| | Effectiveness (F') | | Worst Case | Traceability | Size of |
|---|---|---|---|---|---|
| | Tourism | Finance | Time Complexity | | Hierarchies |
| FCA | **44.69%** | **38.85%** | $O(2^n)$ | **Good** | Large |
| Agglomerative Clustering: | | | | | |
|    Complete Linkage | 36.85% | 33.35% | $O(n^2 \log n)$ | Fair | **Small** |
|    Average Linkage | 36.55% | 32.92% | $O(n^2)$ | | |
|    Single Linkage | 38.57% | 32.15% | $O(n^2)$ | | |
| Bi-Section-KMeans | 36.42% | 32.77% | $O(n^2)$ | Weak | **Small** |

Table 8: Trade-offs between different taxonomy construction methods

Figure 13 shows the number of attributes over the terms' rank, where the rank is a natural number indicating the position of the word in a list ordered by decreasing term frequencies. It can be appreciated that the amount of (non-zero) attributes is distributed in a Zipfian way (compare Zipf, 1932), i.e. a small number of objects have a lot of attributes, while a large number of them has just a few. In particular, for the tourism domain, the term with most attributes is *person* with 3077 attributes, while on average a term has approx. 178 attributes. The total number of attributes considered is 9738, so that we conclude that the object/attribute matrix contains almost 98% zero values. For the finance domain the term with highest rank is *percent* with 2870 attributes, the average being ca. 202 attributes. The total number of attributes is 21542, so that we can state that in this case more than 99% of the matrix is populated with zero-values and thus is much sparser than the ones considered by Lindig (2000). These figures explain why FCA performs efficiently in our experiments. Concluding, though the worst-time complexity is exponential, FCA is much more efficient than the agglomerative clustering algorithm in our setting.

## 7. Related Work

In this section, we discuss some work related to the automatic acquisition of taxonomies. The main paradigms for learning taxonomic relations exploited in the literature are on the one hand clustering

---

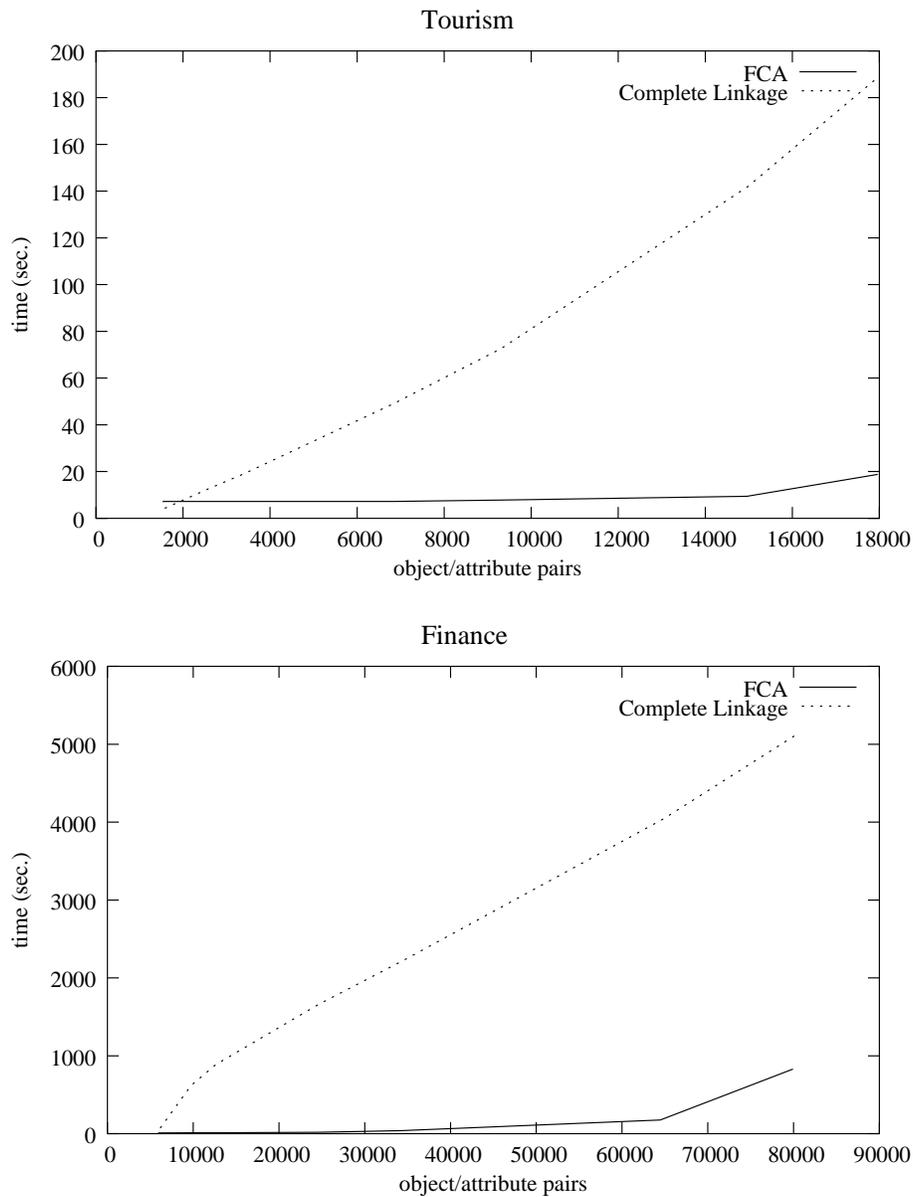

Figure 12: Comparison of the time complexities for FCA and agglomerative clustering for the tourism and finance domains

approaches based on the distributional hypothesis (Harris, 1968) and on the other hand approaches based on matching lexico-syntactic patterns in a corpus which convey a certain relation.

One of the first works on clustering terms was the one by Hindle (1990), in which nouns are grouped into classes according to the extent to which they appear in similar verb frames. In particular, he uses verbs for which the nouns appear as subjects or objects as contextual attributes. Further, he also introduces the notion of *reciprocal similarity*, which is equivalent to our *mutual similarity*. Pereira et al. (1993) also present a top-down clustering approach to build an unlabeled hierarchy of nouns. They present an entropy-based evaluation of their approach, but also show results on a linguistic decision task: i.e. which of two verbs $v$ and $v'$ is more likely to take a given noun $n$ as object. Grefenstette has also addressed the automatic construction of thesauri (Grefenstette, 1994). He presents results on different and various domains. Further, he also compares window-based and syntactic approaches, finding out that the results depend on the frequency of the words in question. In particular, he shows that for frequent words, the syntactic-based approaches are better, while for rare words the window-based approaches are preferable (Grefenstette, 1992). The work of Faure and Nédellec (1998) is also based on the distributional hypothesis; they present an iterative bottom-up clustering approach of nouns appearing in similar contexts. In each step, they cluster the two most similar extents of some argument position of two verbs. Interestingly, this way they not only yield a concept hierarchy, but also ontologically generalized subcategorization frames for verbs. Their method is semi-automatic in that it involves users in the validation of the clusters created in each step. The authors present the results of their system in terms of cluster accuracy in dependency of percentage of the corpus used. Caraballo (1999) also uses clustering methods to derive an unlabeled hierarchy of nouns by using data about conjunctions of nouns and appositions collected from the Wall Street Journal corpus. Interestingly, in a second step she also labels the abstract concepts of the hierarchy by considering the Hearst patterns (see below) in which the children of the concept in question appear as hyponyms. The most frequent hypernym is then chosen in order to label the concept. At a further step she also compresses the produced ontological tree by eliminating internal nodes without a label. The final ontological tree is then evaluated by presenting a random choice of clusters and the corresponding hypernym to three human judges for validation. Bisson et al. (2000) present an interesting framework and a corresponding workbench - Mo'K - allowing users to design conceptual clustering methods to assist them in an ontology building task. In particular they use bottom-up clustering and compare different similarity measures as well as different pruning parameters.

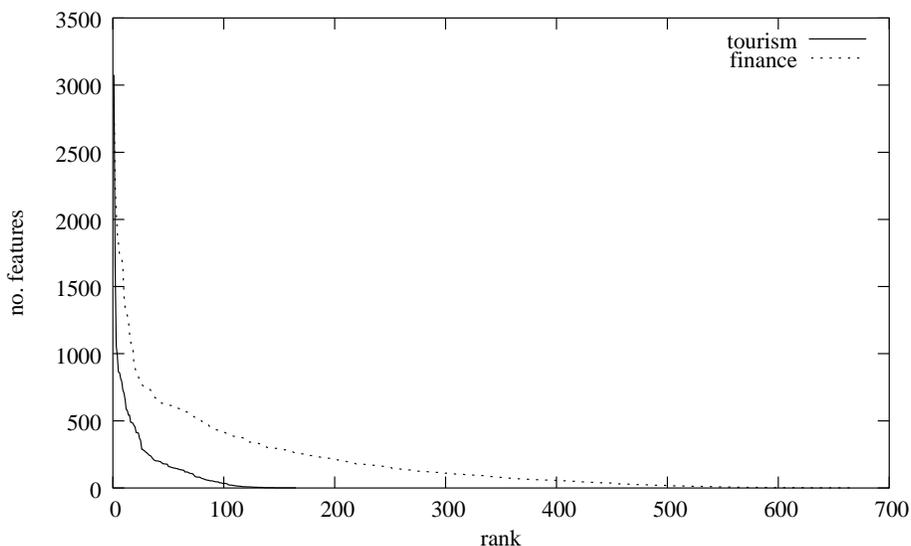

Figure 13: Distribution of Features: number of (non-zero) features over word rank

In earlier work we used collocation statistics to learn relations between terms using a modification of the association rules extraction algorithm (Maedche & Staab, 2000). However, these relations were not inherently taxonomic such that the work described in this paper can not be di-rectly compared to it. Maedche, Pekar, and Staab (2002) examined different supervised techniques based on collocations to find the appropriate hypernym for an unknown term, reaching an accuracy of around 15% using a combination of a tree ascending algorithm and $k$-Nearest-Neighbors as well as the Skew Divergence as similarity measure. These results are neither comparable to the task at hand. Recently, Reinberger and Spyns (2005) have presented an application of clustering techniques in the biomedical domain. They evaluate their clusters by directly comparing to the UMLS thesaurus. Their results are very low (3-17% precision depending on the corpus and clustering technique) and comparable to the results we obtained when comparing our clusters directly with our gold standards and which are not reported in this paper though.

Furthermore, there is quite a lot of work related to the use of linguistic patterns to discover certain ontological relations from text. Hearst's seminal approach aimed at discovering taxonomic relations from electronic dictionaries (Hearst, 1992). The precision of the *isa*-relations learned is $61/106$ (57.55%) when measured against WordNet as gold standard. Hearst's idea has been reapplied by different researchers with either slight variations in the patterns used (Iwanska et al., 2000), in very specific domains (Ahmad et al., 2003), to acquire knowledge for anaphora resolution (Poesio, Ishikawa, im Walde, & Viera, 2002), or to discover other kinds of semantic relations such as part-of relations (Charniak & Berland, 1999) or causation relations (Girju & Moldovan, 2002).

The approaches of Hearst and others are characterized by a (relatively) high precision in the sense that the quality of the learned relations is very high. However, these approaches suffer from a very low recall which is due to the fact that the patterns are very rare. As a possible solution to this problem, in the approach of Cimiano, Pivk, Schmidt-Thieme, and Staab (2004, 2005) Hearst patterns matched in a corpus and on the Web as well as explicit information derived from other resources and heuristics are combined yielding better results compared to considering only one source of evidence on the task of learning superconcept relations. In general, to overcome such data sparseness problems, researchers are more and more resorting to the WWW as for example Markert, Modjeska, and Nissim (2003). In their approach, Hearst patterns are searched for on the WWW by using the Google API in order to acquire background knowledge for anaphora resolution. Agirre, Ansa, Hovy, and Martinez (2000), download related texts from the Web to enrich a given ontology. Cimiano, Handschuh, and Staab (2004a) as well as Cimiano, Ladwig, and Staab (2005) have used the Google API to match Hearst-like patterns on the Web in order to (i) find the best concept for an unknown instance as well as (ii) the appropriate superconcept for a certain concept in a given ontology (Cimiano & Staab, 2004).

Velardi, Fabriani, and Missikoff (2001) present the OntoLearn system which discovers i) the domain concepts relevant for a certain domain, i.e. the relevant terminology, ii) named entities, iii) 'vertical' (is-a or taxonomic) relations as well as iv) certain relations between concepts based on specific syntactic relations. In their approach a 'vertical' relation is established between a term $t_1$ and a term $t_2$, i.e. *is-a*($t_1$,$t_2$), if $t_2$ can be gained out of $t_1$ by stripping of the latter's prenominal modifiers such as adjectives or modifying nouns. Thus, a 'vertical' relation is for example established between the term *international credit card* and the term *credit card*, i.e. *is-a(international credit card,credit card)*. In a further paper (Velardi, Navigli, Cuchiarelli, & Neri, 2005), the main focus is on the task of word sense disambiguation, i.e. of finding the correct sense of a word with respect to a general ontology or lexical database. In particular, they present a novel algorithm called SSI relying on the structure of the general ontology for this purpose. Furthermore, they include an explanation component for users consisting in a gloss generation component which generates definitions for terms which were found relevant in a certain domain.

Sanderson and Croft (1999) describe an interesting approach to automatically derive a hierarchy by considering the document a certain term appears in as context. In particular, they present a document-based definition of subsumption according to which a certain term $t_1$ is more special than a term $t_2$ if $t_2$ also appears in all the documents in which $t_1$ appears.

Formal Concept Analysis can be applied for many tasks within Natural Language Processing. Priss (2004) for example, mentions several possible applications of FCA in analyzing linguistic structures, lexical semantics and lexical tuning. Sporleder (2002) and Petersen (2002) apply FCA to yield more concise lexical inheritance hierarchies with regard to morphological features such as numerus, gender etc. Basili, Pazienza, and Vindigni (1997) apply FCA to the task of learning subcategorization frames from corpora. However, to our knowledge it has not been applied before to the acquisition of domain concept hierarchies such as in the approach presented in this paper.

## 8. Conclusion

We have presented a novel approach to automatically acquire concept hierarchies from domain-specific texts. In addition, we have compared our approach with a hierarchical agglomerative clustering algorithm as well as with Bi-Section-KMeans and found that our approach produces better results on the two datasets considered. We have further examined different information measures to weight the significance of an attribute/object pair and concluded that the conditional probability works well compared to other more elaborate information measures. We have also analyzed the impact of a smoothing technique in order to cope with data sparseness and found that it doesn't improve the results of the FCA-based approach. Further, we have highlighted advantages and disadvantages of the three approaches.

Though our approach is fully automatic, it is important to mention that we do not believe in fully automatic ontology construction without any user involvement. In this sense, in the future we will explore how users can be involved in the process by presenting him/her ontological relations for validation in such way that the necessary user feedback is kept at a minimum. On the other hand, before involving users in a semi-automatic way it is necessary to clarify how good a certain approach works per se. The research presented in this paper has had this aim. Furthermore, we have also proposed a systematic way of evaluating ontologies by comparing them to a certain human-modeled ontology. In this sense our aim has also been to establish a baseline for further research.

## Acknowledgments

We would like to thank all our colleagues for feedback and comments, in particular Gerd Stumme for clarifying our FCA-related questions. We would also like to thank Johanna Völker for comments on a first version as well as for proof-reading the final version of the paper. All errors are of course our own. We would also like to acknowledge the reviewers of the Journal of Artificial Intelligence Research as well as the ones of the earlier workshops (ATEM04, FGML04) and conferences (LREC04, ECAI04) on which this work was presented for valuable comments. Philipp Cimiano is currently supported by the Dot.Kom project (http://www.dot-kom.org), sponsored by the EC as part of the framework V, (grant IST-2001-34038) as well as by the SmartWeb project (http://smartweb.dfki.de), funded by the German Ministry of Education and Research.

## Appendix A. Mutually Similar Terms

| Jaccard | Cosine | L1 norm | Jensen-Shannon divergence |
|---|---|---|---|
| (art exhibition,thing) | (agreement,contract) | (day,time) | (group,person) |
| (autumn,spring) | (animal,plant) | (golf course,promenade) | |
| (balcony,menu) | (art exhibition,washing machine) | (group,person) | |
| (ballroom,theatre) | (basilica,hair dryer) | | |
| (banquet,ship) | (boat,ship) | | |
| (bar,pub) | (cabaret,email) | | |
| (basilica,hair dryer) | (cheque,pension) | | |
| (beach,swimming pool) | (city,town) | | |
| (billiard,sauna) | (conference room,volleyball field) | | |
| (bus,car) | (golf course,promenade) | | |
| (caravan,tree) | (group,party) | | |
| (casino,date) | (inn,yacht) | | |
| (cinema,fitness studio) | (journey,meal) | | |
| (city,town) | (kiosk,tennis court) | | |
| (conference,seminar) | (law,view) | | |
| (conference room,volleyball field) | (library,museum) | | |
| (cure,washing machine) | (money,thing) | | |
| (day tour,place) | (motel,port) | | |
| (distance,radio) | (pilgrimage,whirlpool) | | |
| (exhibition,price list) | (sauna,swimming) | | |
| (ferry,telephone) | | | |
| (gallery,shop) | | | |
| (golf course,promenade) | | | |
| (holiday,service) | | | |
| (journey,terrace) | | | |
| (kiosk,time interval) | | | |
| (law,presentation) | | | |
| (lounge,park) | | | |
| (motel,port) | | | |
| (nature reserve,parking lot) | | | |
| (night,tourist) | | | |
| (region,situation) | | | |

Table 9: Mutually Similar Terms for the tourism domain

| Jaccard | Cosine | L1 norm | Jensen-Shannon divergence |
|---|---|---|---|
| (action,average) | (access,advantage) | (archives,futures) | (cent,point) |
| (activity,downturn) | (acquisition,merger) | (assurance,telephone number) | (government,person) |
| (addition,liquidity) | (action,measure) | (balancing,countenance) | (month,year) |
| (afternoon,key) | (administration costs,treasury stock) | (cent,point) | |
| (agency,purchase) | (advice,assurance) | (creation,experience) | |
| (agreement,push) | (allocation,length) | (government,person) | |
| (alliance,project team) | (amount,total) | (loss,profit) | |
| (allocation,success) | (analysis,component) | (month,year) | |
| (analysis,negotiation) | (area,region) | | |
| (animal,basis) | (arrangement,regime) | | |
| (anomaly,regression) | (assembly,chamber) | | |
| (archives,futures) | (assessment,receipt) | | |
| (area,profitability) | (backer,gamble) | | |
| (argument,dismantling) | (balancing,matrix) | | |
| (arrangement,capital market) | (bank,company) | | |
| (arranger,update) | (barometer,market price) | | |
| (assembly,price decline) | (bid,offer) | | |
| (assurance,telephone number) | (bond,stock) | | |
| (automobile,oil) | (bonus share,cassette) | | |
| (backer,trade partner) | (boom,turnaround) | | |
| (balance sheet,person) | (bull market,tool) | | |
| (balancing,countenance) | (business deal,graph) | | |
| (behaviour,business partnership) | (buy,stop) | | |
| (bike,moment) | (capital stock,profit distribution) | | |
| (billing,grade) | (caravan,software company) | | |
| (board,spectrum) | (cent,point) | | |
| (board chairman,statement) | (change,increase) | | |
| (bonus,nationality) | (commission,committee) | | |
| (bonus share,cassette) | (company profile,intangible) | | |
| (branch office,size) | (complaint,request) | | |
| (broker,competition) | (controller,designer) | | |
| (budget,regulation) | (copper,share index) | | |
| (builder,devices) | (copy,push) | | |
| (building,vehicle) | (credit,loan) | | |
| (business volume,outlook) | (credit agreement,credit line) | | |
| (business year,quota) | (currency,dollar) | | |
| (capital,material costs) | (decision,plan) | | |
| (capital increase,stock split) | (detail,test) | | |
| (capital stock,profit distribution) | (diagram,support) | | |
| (caravan,seminar) | (dimension,surcharge) | | |
| (cent,point) | (discussion,negotiation) | | |
| (chance,hope) | (diversification,milestone) | | |
| (change,subsidiary) | (do,email) | | |
| (charge,suspicion) | (document,letter) | | |
| (chip,woman) | (effect,impact) | | |
| (circle,direction) | (equity fund,origin) | | |
| (clock,ratio) | (evaluation,examination) | | |
| (code,insurance company) | (example,hint) | | |
| (comment,foundation) | (first,meter) | | |
| (commission,expansion) | (forecast,stock market activity) | | |
| (communication,radio) | (function,profile) | | |
| (community,radius) | (gesture,input) | | |
| (company profile,intangible) | (guarantee,solution) | | |
| (compensation,participation) | (half,quarter) | | |
| (complaint,petition) | (increment,rearrangement) | | |
| (computer,cooperation) | (information,trading company) | | |
| (conference,height) | (insurance,percentage) | | |
| (confidentiality,dollar) | (interest rate,tariff) | | |
| (consultant,survey) | (man,woman) | | |
| (contact,hint) | (maximum,supervision) | | |
| (contract,copyright) | (meeting,talk) | | |
| (control,data center) | (merchant,perspective) | | |
| (conversation,output) | (month,week) | | |
| (copper,replacement) | (press conference,seminar) | | |
| (corporation,liabilities) | (price,rate) | | |
| (cost,equity capital) | (productivity,traffic) | | |
| (course,step) | (profit,volume) | | |
| (court,district court) | (share price,stock market) | | |
| (credit,disbursement) | (stock broker,theory) | | |
| (credit agreement,overview) | | | |
| (currency,faith) | | | |
| (curve,graph) | | | |
| (decision,maximum) | | | |
| (deficit,negative) | | | |
| (diagram,support) | | | |
| (difference,elimination) | | | |

Table 10: Mutually Similar Terms for the finance domain

| Jaccard | Cosine | L1 norm | Jensen-Shannon divergence |
|---|---|---|---|
| (disability insurance,pension) | | | |
| (discrimination,union) | | | |
| (diversification,request) | | | |
| (do,email) | | | |
| (effect,help) | | | |
| (employer,insurance) | | | |
| (energy,rest) | | | |
| (equity fund,origin) | | | |
| (evening,purpose) | | | |
| (event,manager) | | | |
| (examination,registration) | | | |
| (example,source) | | | |
| (exchange,volume) | | | |
| (exchange risk,interest rate) | | | |
| (experience,questionnaire) | | | |
| (expertise,period) | | | |
| (faculty,sales contract) | | | |
| (fair,product) | | | |
| (flop,type) | | | |
| (forecast,stock market activity) | | | |
| (fusion,profit zone) | | | |
| (gamble,thing) | | | |
| (good,service) | | | |
| (government bond,life insurance) | | | |
| (happiness,question) | | | |
| (hold,shareholder) | | | |
| (hour,pay) | | | |
| (house,model) | | | |
| (idea,solution) | | | |
| (impact,matter) | | | |
| (improvement,situation) | | | |
| (index,wholesale) | | | |
| (information,trading company) | | | |
| (initiation,middle) | | | |
| (input,traffic) | | | |
| (institute,organization) | | | |
| (investment,productivity) | | | |
| (knowledge,tradition) | | | |
| (label,title) | | | |
| (letter,reception) | | | |
| (level,video) | | | |
| (license,reward) | | | |
| (loan,project) | | | |
| (location,process) | | | |
| (loss,profit) | | | |
| (man,trainee) | | | |
| (margin,software company) | | | |
| (market,warranty) | | | |
| (market access,name) | | | |
| (matrix,newspaper) | | | |
| (meeting,oscillation) | | | |
| (meter,share) | | | |
| (method,technology) | | | |
| (milestone,state) | | | |
| (month,year) | | | |
| (mouse,option) | | | |
| (multiplication,transfer) | | | |
| (noon,press conference) | | | |
| (occasion,talk) | | | |
| (opinion,rivalry) | | | |
| (personnel,resource) | | | |
| (picture,surcharge) | | | |
| (plane,tool) | | | |
| (police,punishment) | | | |
| (profession,writer) | | | |
| (property,qualification) | | | |
| (provision,revenue) | | | |
| (requirement,rule) | | | |
| (risk,trust) | | | |
| (sales revenue,validity) | | | |
| (savings bank,time) | | | |
| (segment,series) | | | |
| (show,team) | | | |
| (speech,winter) | | | |
| (stock broker,theory) | | | |
| (supplier,train) | | | |
| (tariff,treasury stock) | | | |
| (weekend,wisdom) | | | |

Table 11: Mutually Similar Terms for the finance domain (Cont'd)